\pdfoutput=1

\documentclass[11pt]{article}

\usepackage[final]{acl}

\usepackage{times}
\usepackage{latexsym}

\usepackage[T1]{fontenc}

\usepackage[utf8]{inputenc}

\usepackage{microtype}

\usepackage{inconsolata}

\usepackage{graphicx,subcaption,xcolor}
\usepackage{multirow}

\usepackage{amsmath}

\title{ 
LLMs' 
morphological analyses of
complex FST-generated Finnish
words
}

\author{Anssi Moisio$^1$, Mathias Creutz$^2$, \and Mikko Kurimo$^1$ \\
        $^1$Department of Information and Communications Engineering, Aalto University, Finland \\
        $^2$Department of Digital Humanities, University of Helsinki, Finland \\
        \texttt{anssi.moisio@aalto.fi}, \texttt{mathias.creutz@helsinki.fi}, \texttt{mikko.kurimo@aalto.fi}}

\begin{document}
\maketitle
\begin{abstract}

Rule-based language processing systems have been overshadowed by neural systems in terms of utility, but it remains unclear whether neural NLP systems, in practice, learn the grammar rules that humans use.
This work aims to shed light on the issue by evaluating 
state-of-the-art LLMs in a task of morphological analysis of complex Finnish noun forms. We generate the forms using an FST tool, and they are unlikely to have occurred in the training sets of the LLMs, therefore requiring morphological generalisation capacity.
We find that GPT-4-turbo has some difficulties in the task while GPT-3.5-turbo struggles and smaller models Llama2-70B and Poro-34B fail nearly completely.
\end{abstract}

\section{Do neural networks learn grammar?} \label{sec:intro} 

The
debate on whether neural networks (NNs) can be accurate models of human language often revolves around the question whether NNs learn similar grammar rules as children do. In a famous instance of the debate,
\citet{rumelhart1986learning} argued that a NN can capture the implicit rules that govern how English verbs are inflected in the past tense. In a response, \citet{pinker1988language} counter that explicit rules are indispensable to explain how children learn past tenses, and more generally to explain the psychology of language.

Neural methods have gradually become more capable of modelling varied aspects of language, which could be viewed as supporting the implicit rules argument. (For updates on the past-tense debate see \citet{kirov-cotterell-2018-recurrent,
corkery-etal-2019-yet,fukatsu-etal-2024-learning-bidirectional}.)
The most recent instances of the debate are over large language models (LLMs), whose language-generation and task-solving capabilities have surprised many. The recent debate consequently concerns modelling human language more generally instead of focusing on specific phenomena such as verb inflection. Considering the success of LLMs, it is clear that they learn some implicit rule-abiding behaviour that enables them to process and generate language competently, but it is still not clear if they learn grammar similarly to humans, or if they learn and employ some other set of rules.

Assessing grammatical knowledge learned by NNs is not straightforward, but there are at least two popular approaches.
Training a classifier (called a `probe' \citep{alain2016understanding} or a `diagnostic classifier' \citep{hupkes2018visualisation}, first developed by \citet{shi-etal-2016-string,adi2017fine}) to classify the internal representations of NNs has been used to inspect what aspects of grammar are encoded in them.
Probing studies have found various syntactical information encoded in neural NLP systems \citep{jawahar-etal-2019-bert,tenney2018you,papadimitriou-etal-2021-deep}, but interpreting the results remains contentious
\citep{voita-titov-2020-information,immer-etal-2022-probing}.

The other popular method
is to directly inspect a neural LM's next-unit predictions, or to train a classifier NN to predict which word is most acceptable, given sequence of previous words.
In an influential work by \citet{linzen-etal-2016-assessing}, knowledge of subject-verb agreement in LSTM networks was assessed this way, and it was concluded that `LSTMs can learn to approximate structure-sensitive dependencies fairly well'.
Similar \emph{targeted syntactic evaluation} methods, inspired by methods in psycholinguistics (e.g. \citet{crain1985can,stowe1986parsing}), have subsequently been employed to assess the knowledge of many different grammatical phenomena in NNs, for example anaphora or negative polarity items  \citep{marvin-linzen-2018-targeted, futrell-etal-2019-neural, jumelet-hupkes-2018-language,
hu-etal-2020-systematic}. Larger test suites
such as BLiMP \citep{warstadt-etal-2020-blimp} or SyntaxGym \citep{gauthier-etal-2020-syntaxgym}
are used as benchmarks to track advances in the field.

The general conclusion has not changed much since that of \citeauthor{linzen-etal-2016-assessing}'s: the networks are \emph{fairly} good at acquiring the grammar rules. Sometimes results of a single study are interpreted as evidence that the NNs have acquired a syntactical rule completely (e.g. \citet{wilcox2023using}), but a closer inspection often proves such an interpretation premature (e.g. \citet{lan2024large}).
Since there is no conclusive evidence that NNs learn from text the same grammar that people use,
it remains an important task to delineate the instances where NNs, and LLMs in particular, adhere to and utilise grammar, and the instances where they do not.

Designing targeted syntactic evaluation tests requires careful formulation of the sequences. For example, \citet{wilcox2023using} examined the understanding of filler-gap effects by comparing the probabilities of acceptable and unacceptable continuations for sentence pairs such as `I know \emph{what} the lion devoured' and `I know \emph{that} the lion devoured'. The continuation `yesterday' is assumed to be acceptable for the former but not the latter sequence.
However, `yesterday' could be an acceptable next word even for the latter sequence: consider the sentence `I know that the lion devoured yesterday's leftovers.' This example highlights the difficulty of designing test sentences of this sort.

Instead of inspecting the next-unit predictions or training diagnostic classifiers, in this work we ask LLMs explicitly to perform a classification task, which is possible due to the flexible text generation capacity of the LLMs.
This
makes the evaluation relatively unambiguous. For example, asking an LLM directly `Is the verb ``devour'' transitive or intransitive?' does not leave much room for confounding factors.
The apparent limitation of this method is that even if a model fails in an explicit classification task like this, we cannot rule out the possibility that the model nevertheless encodes perfect \emph{implicit} knowledge of the verb and how to use it in any context.
However, we make the assumption in this work that if the LLMs had learned a grammar rule as perfectly as humans, they would be able to answer the explicit questions as competently as humans. This seems justified considering the type and difficulty of, and LLMs' performance in, other tasks used to evaluate LLMs, such as academic and professional exams \citep{openai2023gpt}.
%

This approach was also taken by \citet{weissweiler-etal-2023-counting}, who
assessed the morphological competence of GPT-3.5-turbo by asking it directly to fill in past tenses of words in a sentence, and concluded that it `massively underperforms purpose-built systems'. Similarly, \citet{weller-di-marco-fraser-2024-analyzing-understanding} took a morphologically complex word $W$ and asked GPT-3.5-turbo questions such as `What is the head noun of $W$?'.

In this work we present LLMs directly and explicitly with a classification task
to investigate the knowledge of Finnish morphology in LLMs.
Although Finnish has relatively few speakers worldwide (<10 million), it is not a low-resource language, having about 32B tokens of available training texts \citep{luukkonen-etal-2023-fingpt, luukkonen2024poro}. Consequently, the state-of-the-art (SOTA) multilingual LLMs such as GPT-4 are fluent in Finnish, and could be expected to have a good grasp of the grammar, if the LLMs are in fact good at learning grammar from text.



\section{Data and methods}

Previous datasets of inflected Finnish words include the MorphyNet \citep{batsuren-etal-2021-morphynet} and UniMorph \citep{kirov-etal-2016-large,
batsuren-etal-2022-unimorph-short} corpora. We chose not to use data from these datasets for two reasons. Firstly, complex words comprising unusually many morphemes make it possible to assess if the systems can generalise to many types of possible inflections instead of learning only the most common inflection types. The previous datasets do not include many extremely complex word forms, but these can be generated using a finite-state transducer (FST). Secondly, since the SOTA LLMs have been trained on very large datasets harvested from the Internet, it is likely that the previously published datasets are included in their training data, which would preclude fair assessment.

We use the Omorfi tools \cite{pirinen2015development,pirinen2017open} that are based on finite-state morphology \citep{koskenniemi1984general, beesley2003finite} to generate inflected forms of Finnish nouns. The Omorfi library includes some 500k lexemes, of which about 140k are nouns. We inflect the nouns in all possible combinations of number, grammatical case, and possessive suffix (see Table~\ref{tab:morph} for examples, and Appendix~\ref{sec:expt_details_task} for further details),
\begin{table}[htb]
\begin{center}
\small
\begin{tabular}{llll}
    \texttt{BASE}    & \texttt{+PL}      & \texttt{+INE}          & \texttt{+SG2 / +PL1} \\
    \multirow{3}{*}{\textit{laite}}   & \multirow{3}{*}{\textit{laitteet}} & \emph{laitteissa}    & \textit{laitteissasi} / \textit{laitteissamme} \\
            &          & \texttt{+TRA}          & \\
            &          & \textit{laitteiksi}    & \textit{laitteiksesi} / \textit{laitteiksemme} \\
\end{tabular}
\end{center}
\caption{Examples of inflections of the word `laite' (`device'). PL means plural, INE and TRA are case classes, and SG2/PL1 are possessive suffixes. Inflections in each column include also those in the columns to their left.}  \label{tab:morph}
\end{table}
which creates about 25M word forms. A random sample of 2000 inflected nouns is used as a test set in our experiments. We are unaware of any assessment of the generation accuracy of Omorfi, so we performed manual evaluation of the first 200 words in the sample and found 6 incorrectly inflected words. We therefore estimate the generation accuracy to be around 97\%, which creates an upper bound for the classification accuracy of the test set. We publish the test set and all code to reproduce the results at \url{https://github.com/aalto-speech/llm-morph-tests}. We note, however, that once the data is published, it is subject to the same data contamination issue as the previous datasets mentioned above---the good thing is that one can always draw a new random sample from the full set of 25M forms.

Uniform sampling of lexemes creates a bias towards low-frequency types that are correlated with regularity of the inflection \citep{kodner-etal-2023-morphological}. We note that this is the case in our data, as we took a random sample of the lexemes, and this should be kept in mind when interpreting the results; there are probably not many irregularly inflected words, which makes the task easier. This is not an issue, however, given our research question of whether the LLMs have picked up even the most \emph{systematic} inflection types from textual data.

\begin{table}[htb]
\begin{center}
\fontsize{8}{9.5}\selectfont
\begin{tabular}{l}
    \small \textbf{Prompt:} \\ \hline
    
Jäsennä taivutetut substantiivit tällä tavalla: \\
taivutusmuoto -- perusmuoto, luku, sijamuoto, omistusliite \\
 \\
vedessämme -- vesi, yksikkö, inessiivi, 1. persoonan monikko \\
kinoksiksensa -- kinos, monikko, translatiivi, 3. persoona \\
peukalostanne -- peukalo, yksikkö, elatiivi, 2. persoonan monikko \\
huurteenani -- huurre, yksikkö, essiivi, 1. persoonan yksikkö \\
sängiltäsi -- sänki, monikko, ablatiivi, 2. persoonan yksikkö \\
koivuumme -- koivu, yksikkö, illatiivi, 1. persoonan monikko \\
kaistojaan -- kaista, monikko, partitiivi, 3. persoona \\
rehtiyksiesi -- rehtiys, monikko, genetiivi, 2. persoonan yksikkö \\
laaksoillani -- laakso, monikko, adessiivi, 1. persoonan yksikkö \\
talollenne -- talo, yksikkö, allatiivi, 2. persoonan monikko \\
kansoiltanne -- kansa, \\
\hline \\
\small  \textbf{Correct answer:} \\ \hline
monikko, ablatiivi, 2. persoonan monikko \\ \hline
\end{tabular}
\end{center}
\caption{An example 10-shot prompt. An English translation of the first two rows is: \emph{Parse the inflected nouns in this manner: inflected form -- base form, number, grammatical case, possessive suffix}. The following rows are the examples. We use $n$-shot prompts with $n \in \{0,1,5,10\}$, and for all $n$ we use the same $n$ first examples. For instance, the 5-shot prompts have the \emph{vedessämme, kinoksiksensa, peukalostanne, huurteenani}, and \emph{sängiltäsi} example rows.}  \label{tab:prompts}
\end{table}

LLMs are prompted to give a morphological analysis given an inflected form and the base form. That is, the models should give the correct number, case, and possessive suffix classes of the inflected noun. The prompt, shown in Table~\ref{tab:prompts}, comprises a short description of the task and the desired format, after which there are 0, 1, 5, or 10 examples of the task before the test word.


We test \textbf{GPT-4-turbo}-1106-preview \citep{achiam2023gpt} (which outperformed GPT-4-0613 in preliminary experiments), \textbf{GPT-3.5-turbo}-1106, \textbf{Llama2-70B} \citep{touvron2023llama} (outperformed smaller Llama2 models and chat versions), and \textbf{Poro-34B} \citep{luukkonen2024poro}, which is
trained on Finnish, English, and programming code.

For Poro and Llama2, we performed a coarse tuning of the temperature parameter on a validation set, and found no large differences but 0.5 to be marginally better than the others, so we used this value in the experiments with these models.
For the GPT models we found a temperature of 0.0 to yield the best results, so this value is used for GPT-4-turbo and GPT-3.5-turbo. We did not tune the top\_p parameter (of \emph{nucleus sampling}) but used the default value 1.0. 

Additionally, we trained simple recurrent neural network (RNN) models to also classify words (one RNN for each category: number, case, and possessive suffix), using random samples of the FST-generated word forms as training data (excluding the test set). The aim of this comparison is to give some indication of the difficulty of the task, and to see if NNs can handle the task if they are specifically trained on this small subset of Finnish morphology. 
We took the RNN off the shelf of the Pytorch library\footnote{
From the tutorial at
\url{https://pytorch.org/tutorials/intermediate/char_rnn_classification_tutorial}
}
without tuning any of its hyperparameters. It consists of three layers of size 128.

\section{Results}

\begin{figure*}[htb]
  \includegraphics[width=2.0\columnwidth]{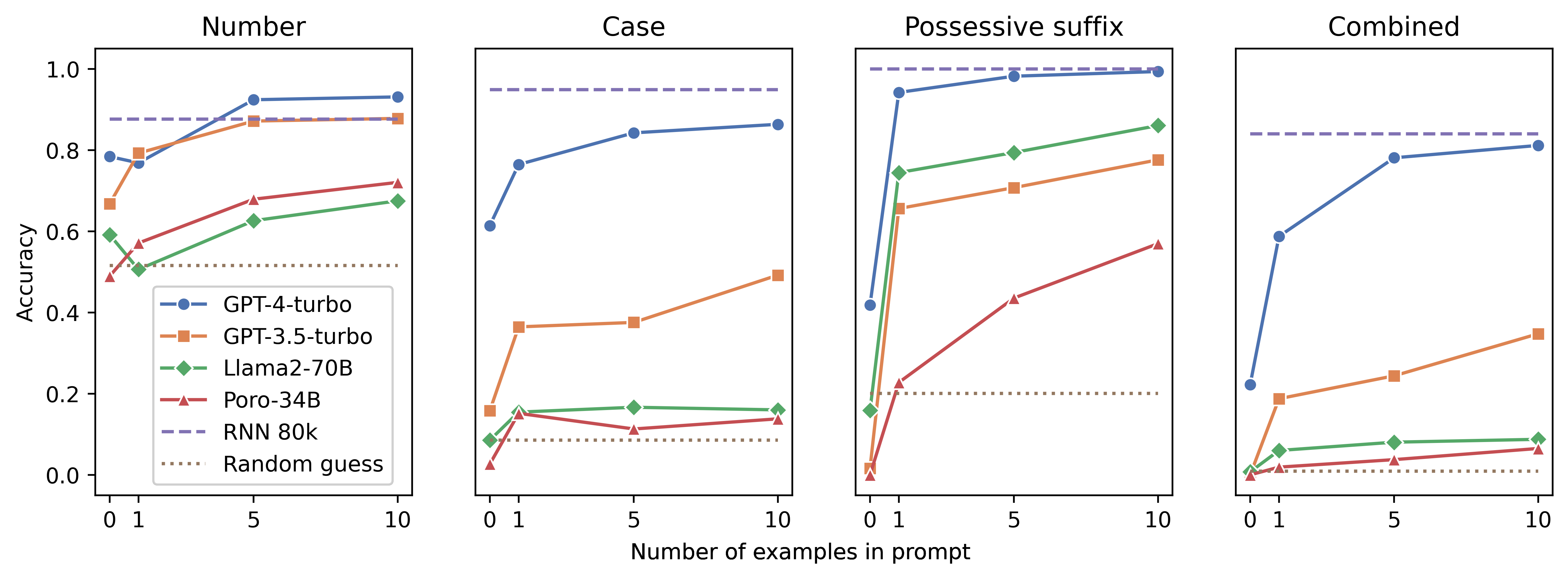}
  \caption{Results in the morphological analysis task.}
  \label{fig:results}
\end{figure*}

The rightmost plot in Figure~\ref{fig:results} shows that besides GPT-4-turbo, the models perform poorly in the task. GPT-4-turbo is not close to perfect accuracy either, and the combined 10-shot result does not reach the result achieved by simple RNNs trained with 80k words. With training set sizes of 800, 4k, 8k, 40k, and 80k words, the RNNs achieved accuracies of 0.380, 0.765, 0.774, 0.821, and 0.840, respectively.

The first three plots from left in Figure~\ref{fig:results} break down the classification task into the three component classification tasks: number, case, and possessive suffix. There are some differences in the strengths of the models: Llama outperforms GPT-3.5 in the possessive suffix classification task, while GPT-3.5 performs better for other classification tasks. In number classification, Poro outperforms Llama, although Llama performs better in other tasks.

\begin{figure}[htb]
    \includegraphics[width=0.99\columnwidth]{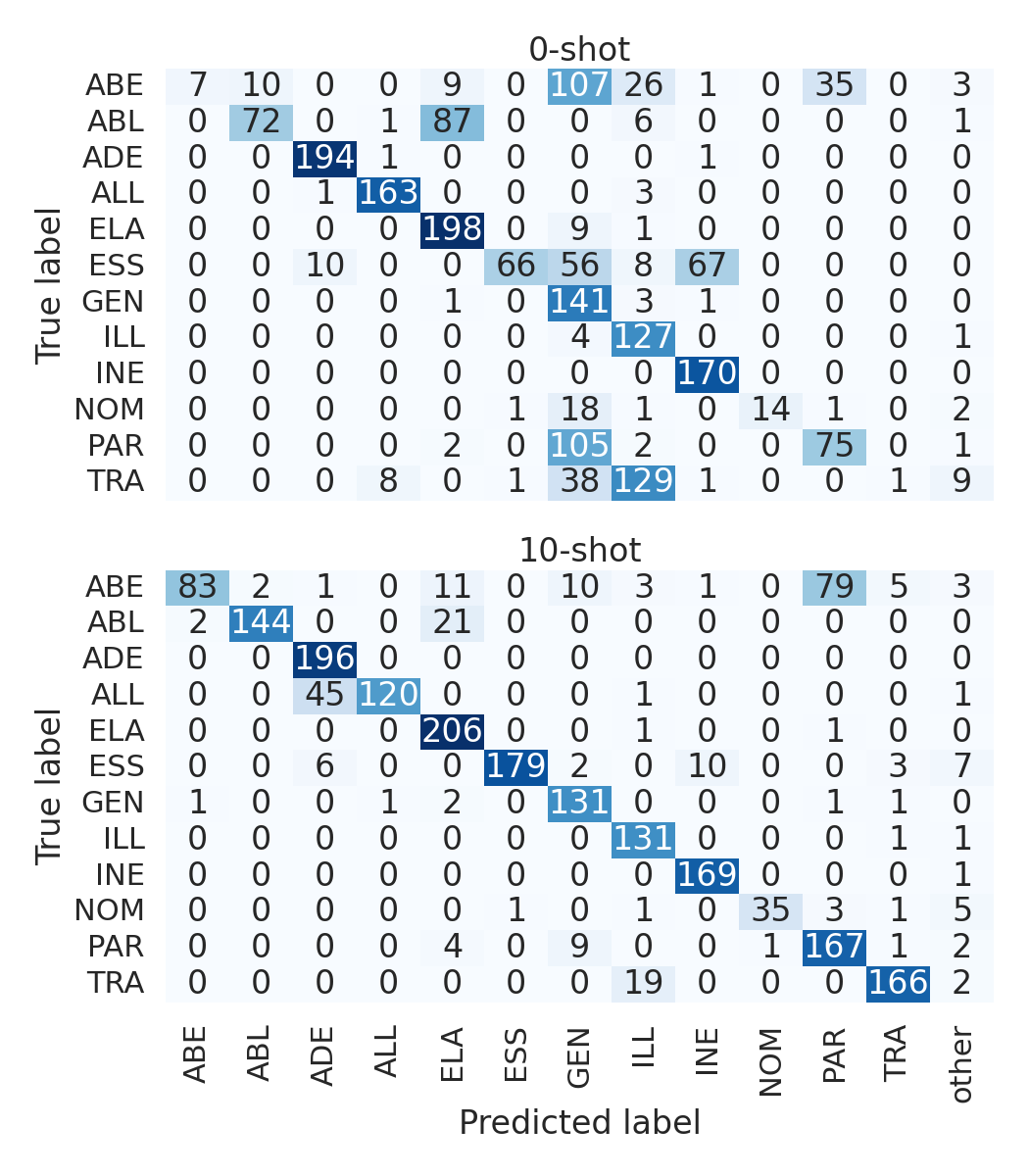}
    \caption{Case label confusions of GPT-4-turbo in the 0-shot and 10-shot setups. See Appendix~\ref{sec:detailed_results} for all confusion matrices.}
    \label{fig:conf-case-gpt4}
\end{figure}

Figure~\ref{fig:conf-case-gpt4} shows the confusion matrices for GPT-4-turbo classifications of cases for the 0-shot and 10-shot setups. From the 0-shot confusion matrix we can see that the model does predict all classes even though we did not provide it with the names of the classes we expected it to recognise. This is not surprising, since GPT-4-turbo has no difficulties if asked to inflect a Finnish word in all cases and to provide the names of the cases. It is obvious that GPT-4-turbo has a fair amount of both
declarative knowledge (metalinguistic knowledge; it knows the classes) and procedural knowledge (knows how to inflect the words)
of the Finnish morphology. Therefore, the challenge in this task comes presumably from the need to generalise to infrequently used, morphologically complex word forms.

\begin{figure}[htb]
    \includegraphics[width=\columnwidth]{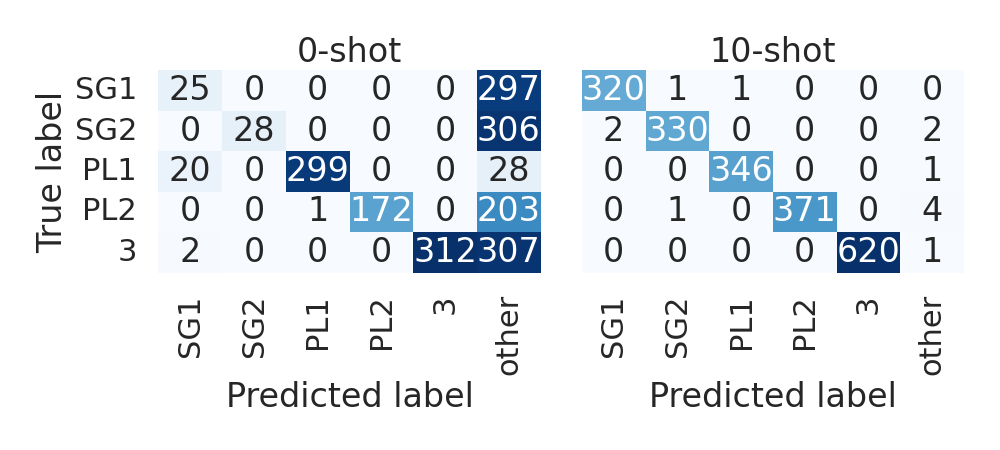}
    \caption{Possessive suffix label confusions of GPT-4-turbo in the 0-shot and 10-shot setups.  See Appendix~\ref{sec:detailed_results} for all confusion matrices.}
    \label{fig:conf-person-gpt4}
\end{figure}

\section{Discussion}

\subsection{Reasons behind the errors}

Most current SOTA LLMs use subword tokenisation methods such as BPE \citep{sennrich-etal-2016-neural} that break down infrequent character sequences into multiple shorter tokens while keeping frequent sequences as single tokens. Intuitively, having long tokens that combine multiple morphemes into a single token could hinder the capacity to model morphology, since multiple embeddings would have to be learned for a single morpheme. Of the three model families, Poro uses the longest tokens, having an average of 3.55 characters per token in our test words, while the Llama average is 2.16 and the GPT average is 2.26. Furthermore, the average length of the last token of a word is even longer: 4.42 for Poro, 2.41 for Llama, and 2.78 for GPT.
For example, the first two test words whose possessive suffix Poro classifies incorrectly and differences in the tokenisations of the different models are shown in Table~\ref{tab:tokenisations}.
\begin{table}[htb]
\begin{center}
\footnotesize
\begin{tabular}{r| ll}
     \textbf{Base form} & \emph{lyhty} (\emph{lantern}) & \emph{tarttuma} (\emph{infection}) \\ 
     \textbf{Test word} & \emph{lyhtyjämme} & \emph{tarttumassamme} \\ 
     \textbf{Poro tokens} & ly hty jämme & t art t um assamme  \\ 
     \textbf{Llama tokens} & ly ht yj äm me & tart t um ass am me \\ 
     \textbf{GPT tokens} & ly ht y j äm me & t art t um ass am me  \\
\end{tabular} \caption{BPE tokenisations of different models.} \label{tab:tokenisations}
\end{center}
\end{table}
Both of these words have the first person plural possessive suffix, which always ends in `me'. The possessive suffix `me' is combined with the case morpheme (partitive `jä' in `lyhtyjämme' and inessive `ssa' in `tarttumassamme')  by Poro but not by GPT or Llama. This might be one reason Poro misclassifies these words, while GPT and Llama do not, and in general why Poro lags behind the other models in the possessive suffix classification task  as seen in Figure~\ref{fig:results}. The possessive suffix is simple to recognise, \emph{if} the tokenisation is conducive to the task: a rule that checks the last two letters of the word and assigns \texttt{`ni'-->SG1; `si'-->SG2; `me'-->PL1; `ne'-->PL2; else-->3} would achieve 100\% accuracy on our test set. Admittedly, the rule would have to be more complicated if there were also words without any possessive suffix, since these words could end in virtually any two letters: for instance, `vesi' (`water') ends in `si' but does not have any possessive suffix (SG2 form would be `vetesi') as does the translative case `vedeksi' without a possessive suffix (the translative case with SG2 suffix becomes `vedeksesi').

Class frequencies could also explain some of the confusions. For example, GPT-4 often confuses abessive cases as partitive, seen in Figure~\ref{fig:conf-case-gpt4}. In addition to partitive being often quite similar to abessive, for example the inflected forms `kättä' and `kädettä' of the base `käsi' (`hand'), partitive is also much more common than abessive: 16.2\% versus 0.1\% of occurrences in \citet{kettunen2005sijamuodot}.




\subsection{Interpretations and implications}
The results suggest that despite the versatile language generation capacity of GPT-4-turbo it has not acquired the rules of Finnish morphology as completely as
could be expected based on its language generation capacity.
Instead, GPT-4 employs some other set of heuristics to decide the next token, although these undoubtedly overlap somewhat with grammar rules. This is hardly a surprise given the literature reviewed in Section~\ref{sec:intro}, where the general conclusion tends to be that NNs rarely use grammar rules systematically, although usually fairly well.


The ineptitude of neural nets to follow grammar rules is related to systematic compositionality and inefficiency w.r.t training data set size, which are said to be weaknesses of neural nets compared to rule-based systems.
%
%
%
Learning grammar enables \emph{systematic compositional generalisation} \citep{fodor1988connectionism}: learning a concise grammar rule such as `\texttt{the suffix -nne indicates 2nd person plural possessive form}' would enable generalising to all possible 2nd person plural forms in Finnish, obviating the need to learn word-specific associations and therefore reducing the required training corpus size. GPT-4 reaches close to 100\% accuracy in this simple task of classifying possessive suffixes (RNN reaches 100\%, and it is obvious that Finnish speakers would also reach 100\%). However, the fact that it still sometimes classifies words ending in `nne' as 2nd person singular instead of plural (see Figure~\ref{fig:conf-person-gpt4}) betrays its incomplete grasp of the systematic possessive suffixes in Finnish. Similar arguments apply to the other two classification tasks and the combined classification task.





\section{Conclusion}
We conclude that even a SOTA LLM, GPT-4-turbo, does not model Finnish morphology
thoroughly enough to allow it to provide morphological analyses of rare and complex word forms with a high accuracy. Contrasting this with its impressive text generation capacity 
suggests that it utilises some other
language processing heuristics, which clearly overlap somewhat with morphological rules since it rarely produces incorrect forms, but which preclude human-level systematic generalisation on our test set. GPT-4-turbo outperforms models such as GPT-3.5-turbo and Llama2-70B, however, by a large margin.

\section{Limitations}

Our experiments are limited to only one language and only four LLMs, which of course means we cannot be certain how the models perform on different languages, or how other models perform in Finnish, even though we suggest our results shed some light on general questions of grammar represented in LLMs. We also have not optimised the prompt beyond trying out a few different phrasings, so we assume some other prompt could elicit better performance especially in the 0- and 1-shot setups.

As noted in the introduction, we assess LLMs using explicit, metalinguistic questions about Finnish morphology. It is in principle possible that even if the models fail in this task, having a limited grasp of the morphological labels, they could succeed in using the words correctly in sentences and representing their meanings correctly.

\section{Acknowledgements}
We thank the anonymous reviewers for their insightful comments and feedback.
The work was supported by the Finnish Cultural Foundation grant 00240853 and the Academy of Finland grant 337073.
The computational resources were provided by Aalto ScienceIT. The use of the GPT-3.5 and GPT-4 systems via the Azure OpenAI API was provided by Aalto IT Services.

\bibliography{anthology,custom}

\appendix


\section{Details of the classification task} \label{sec:expt_details_task}

We inflect Finnish nouns in all possible combinations of number, grammatical case, and possessive suffix. Tables~\ref{tab:cases} and \ref{tab:persons} list the classes of case and possessive suffix with examples of both singular and plural forms. We include a possessive suffix in all the forms in our test set.

\begin{table}[htb]
    \begin{center}
    \begin{tabular}{llll}
     \textbf{Short} &        \textbf{Name}  & \textbf{SG e.g.}        & \textbf{PL e.g.} \\
    \hline
    ABE   &      abessive   &  talotta  & taloitta \\
    ABL   &      ablative   &   talolta & taloilta \\
    ADE   &      adessive   &   talolla & taloilla \\
    ALL   &      allative   &   talolle & taloille \\
    ELA   &      elative    &   talosta & taloista \\
    ESS    &     essive     &   talona  & taloina \\
    GEN    &     genitive   &   talon   & talojen \\
    ILL    &     illative   &   taloon  & taloihin \\
    INE   &      inessive   &  talossa  & taloissa \\
    NOM   &      nominative &   talo    & talot \\
    PAR    &     partitive  &  taloa    & taloja \\
    TRA   &      translative&  taloksi  & taloiksi \\
    \end{tabular}
    \end{center}
    \caption{Finnish grammatical cases used in the experiments, with example inflections of the word `talo' (`house'). There are three more grammatical cases in Finnish (totalling 15), but comitative and instructive are not supported by Omorfi, and accusative does not have its own unambiguous surface form, so these three are not included in our data.} \label{tab:cases}
\end{table}

\begin{table}[htb]
    \begin{center}
    \begin{tabular}{lll}
     \textbf{Class} &   \textbf{SG e.g. (ELA)}        & \textbf{PL e.g. (ELA)} \\
    \hline
    - &  talosta & taloista \\
    SG1         & talostani     & taloistani \\
    SG2         & talostasi     & taloistasi \\
    PL1         & talostamme    & taloistamme \\
    PL2         & talostanne    & taloistanne \\
    \multirow{2}{*}{3}  & talostaan,   & taloistaan,   \\
                        & talostansa   & taloistansa
    \end{tabular}
    \end{center}
    \caption{Possessive suffixes in Finnish, with example inflections of the word `talo' (`house') with the elative grammatical case `talosta'. SG1 is `first person singular', SG2 is `second person singular' etc. The third person has the same forms in singular and plural, but there are synonyms such as `talostaan' and `talostansa'.}  \label{tab:persons}
\end{table}





\section{Detailed results} \label{sec:detailed_results}



\begin{figure}[htb]
    \begin{subfigure}[b]{0.23\textwidth}
      \includegraphics[width=\textwidth]{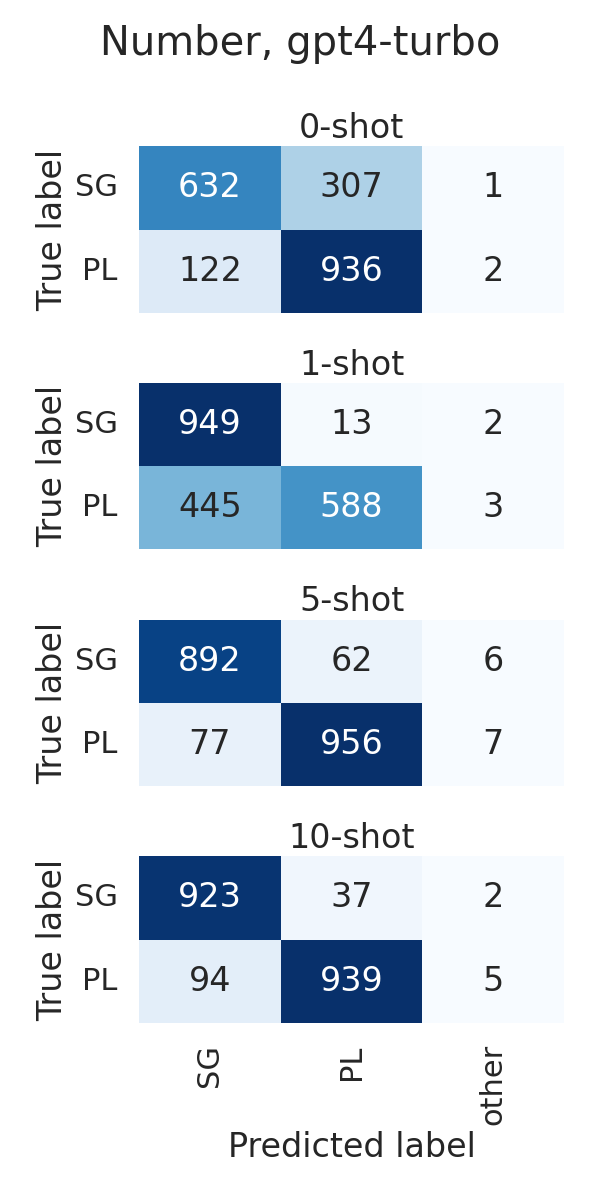}
    \end{subfigure}
    \begin{subfigure}[b]{0.23\textwidth}
      \includegraphics[width=\textwidth]{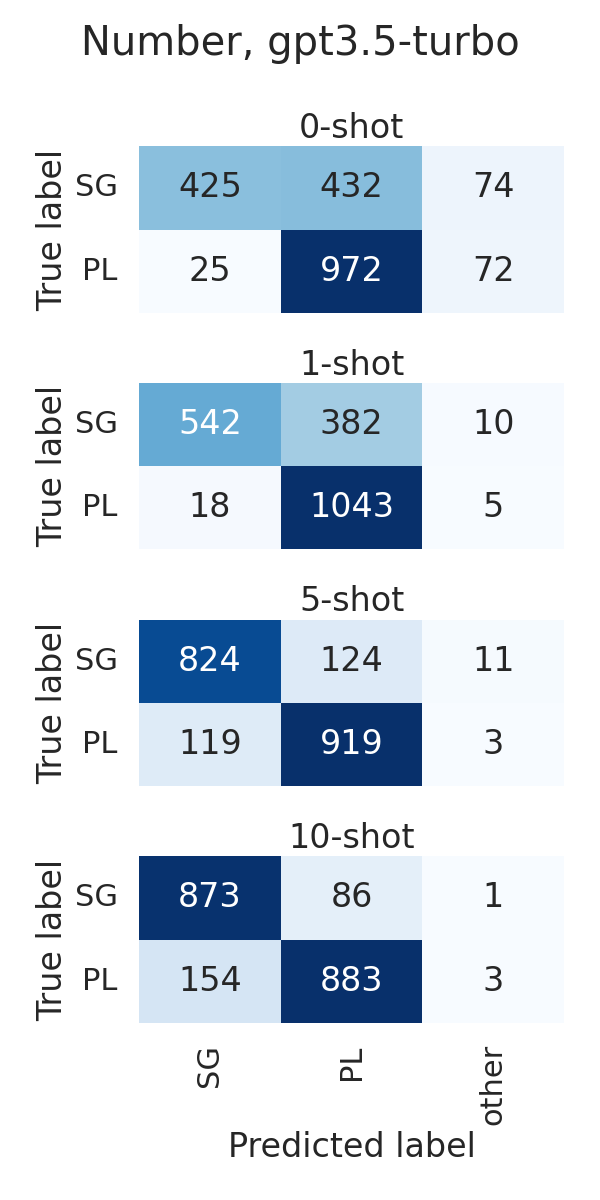}
    \end{subfigure}
    \caption{Confusions in the GPT-4-turbo and GPT-3.5-turbo number classification task.}
    \label{fig:num-gpt}
\end{figure}

\begin{figure}[htb]
    \begin{subfigure}[b]{0.23\textwidth}
      \includegraphics[width=\textwidth]{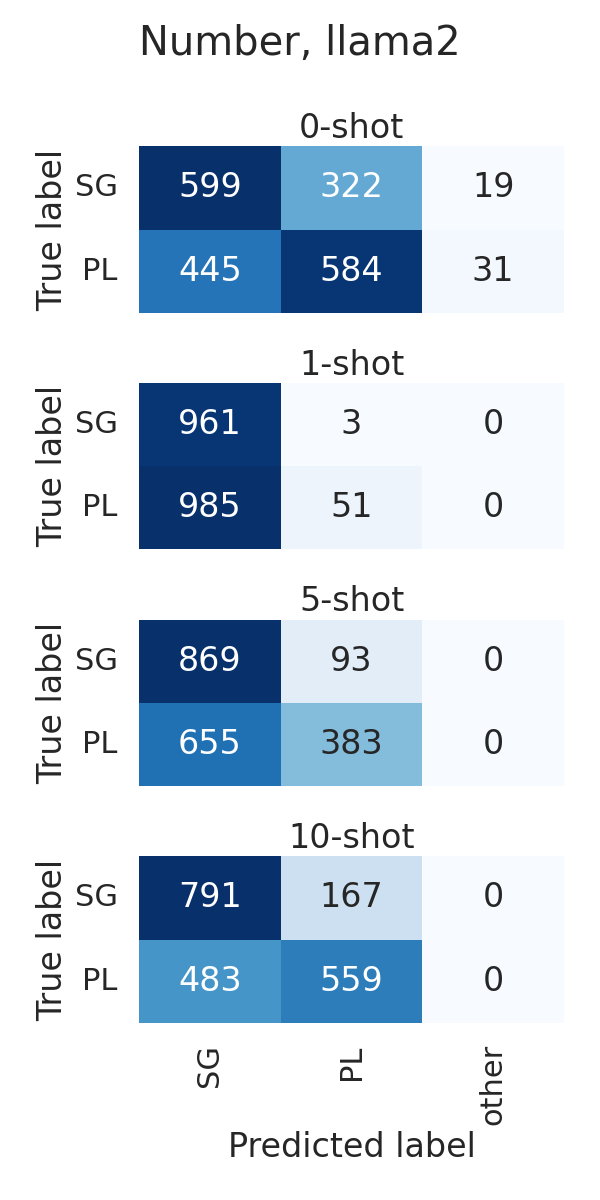}
    \end{subfigure}
    \begin{subfigure}[b]{0.23\textwidth}
      \includegraphics[width=\textwidth]{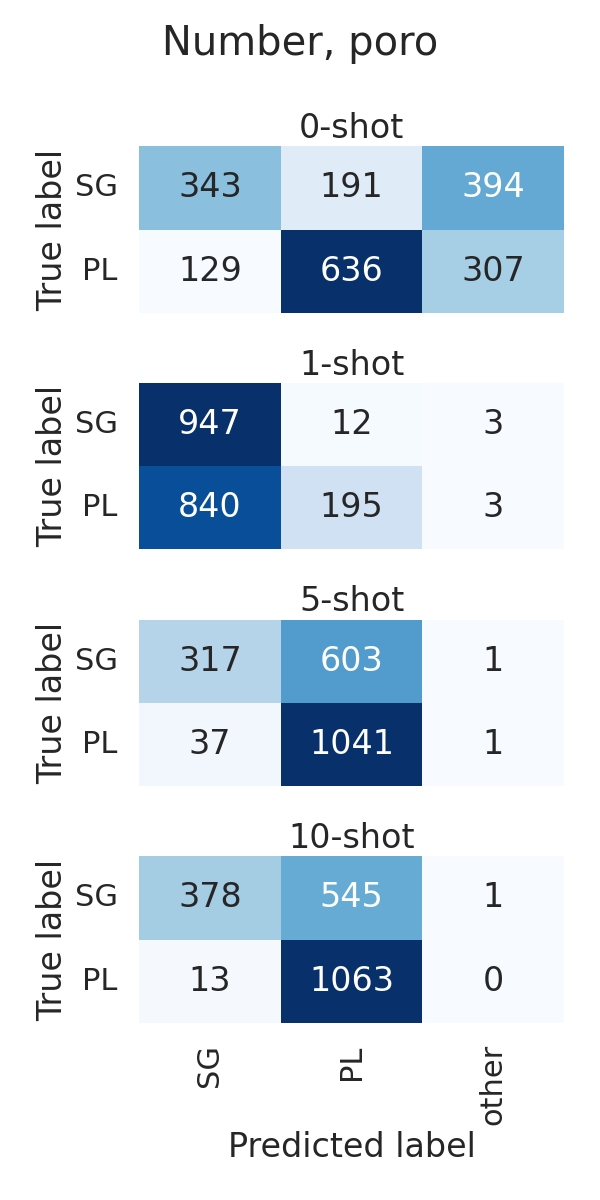}
    \end{subfigure}
    \caption{Confusions in the Llama2-70B and Poro-34B number classification task.}
    \label{fig:num-llama-poro}
\end{figure}

\begin{figure}[htb]
    \includegraphics[width=\columnwidth]{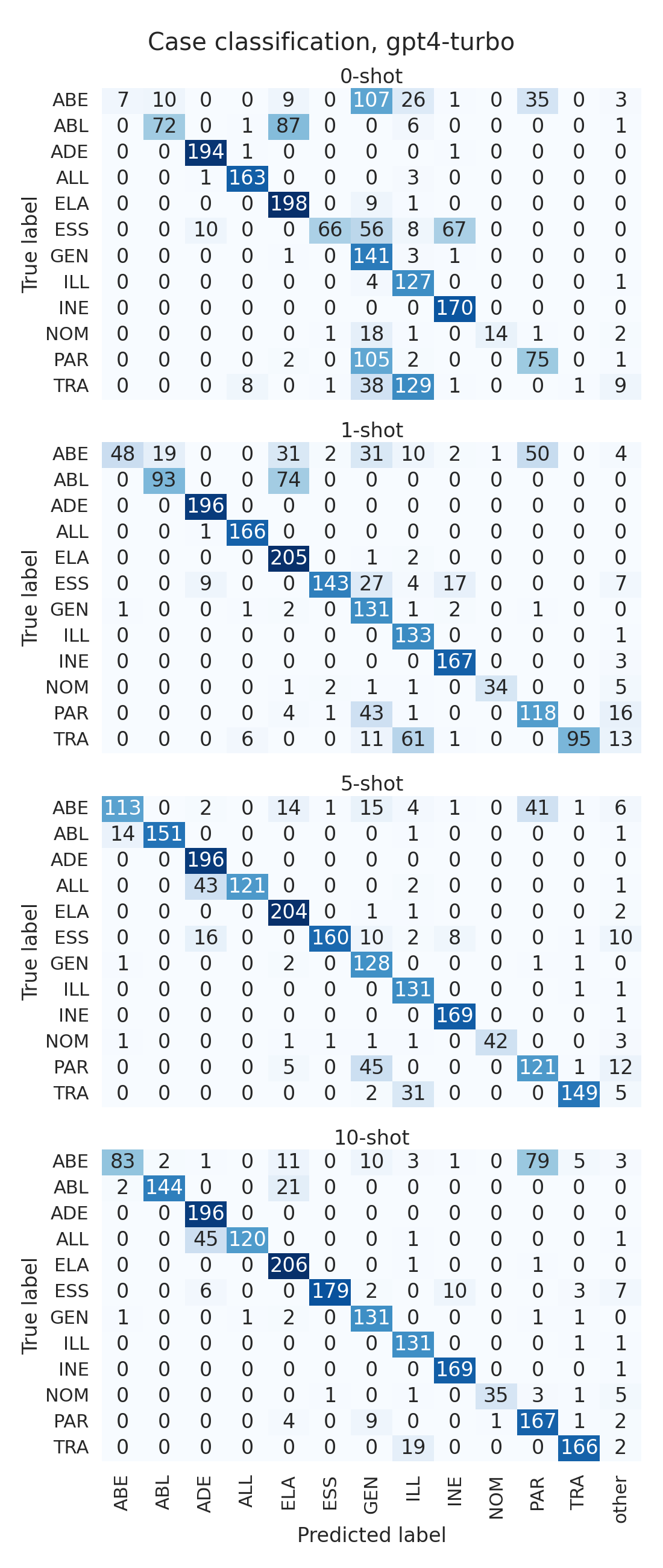}
    \caption{Confusions of GPT-4-turbo in the case classification task.}
    \label{fig:conf-gpt4-case}
\end{figure}

\begin{figure}[htb]
    \includegraphics[width=\columnwidth]{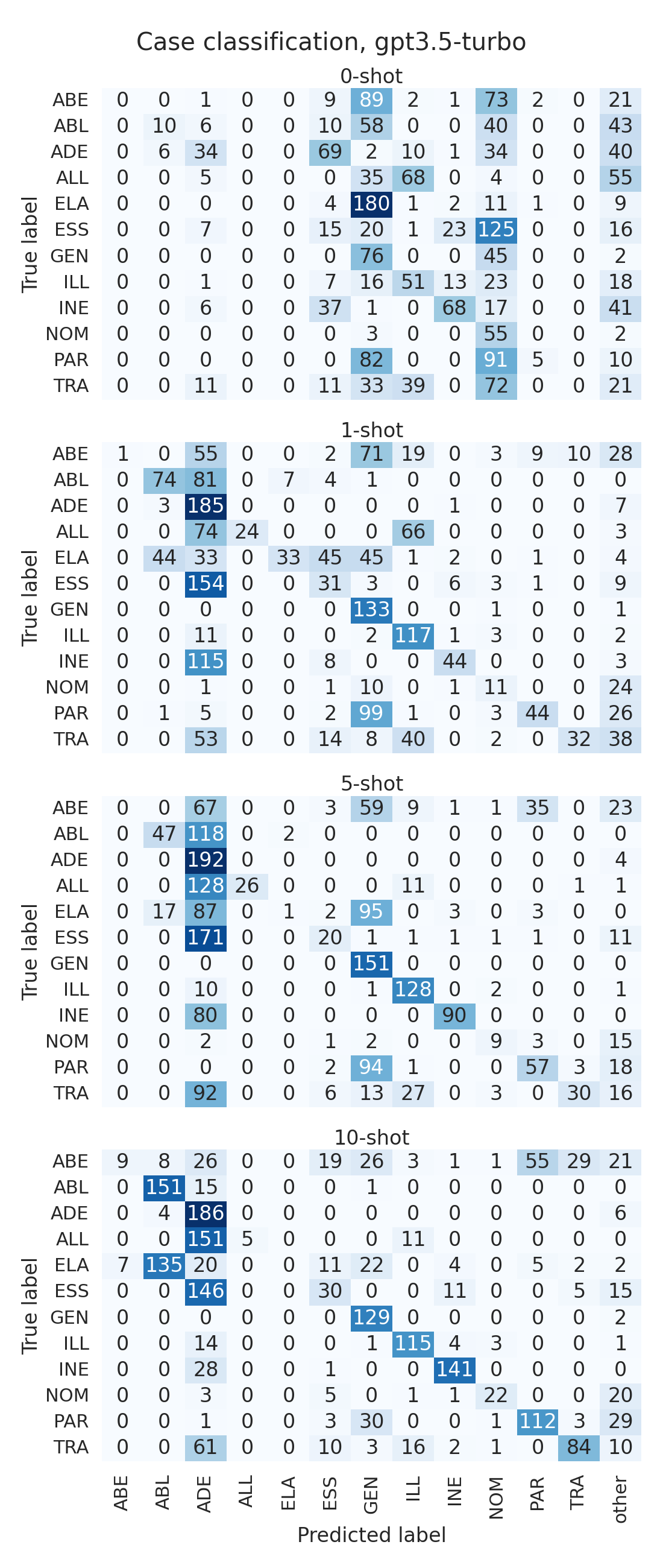}
    \caption{Confusions of GPT-3.5-turbo in the case classification task.}
    \label{fig:conf-gpt35-case}
\end{figure}

\begin{figure}[htb]
    \includegraphics[width=\columnwidth]{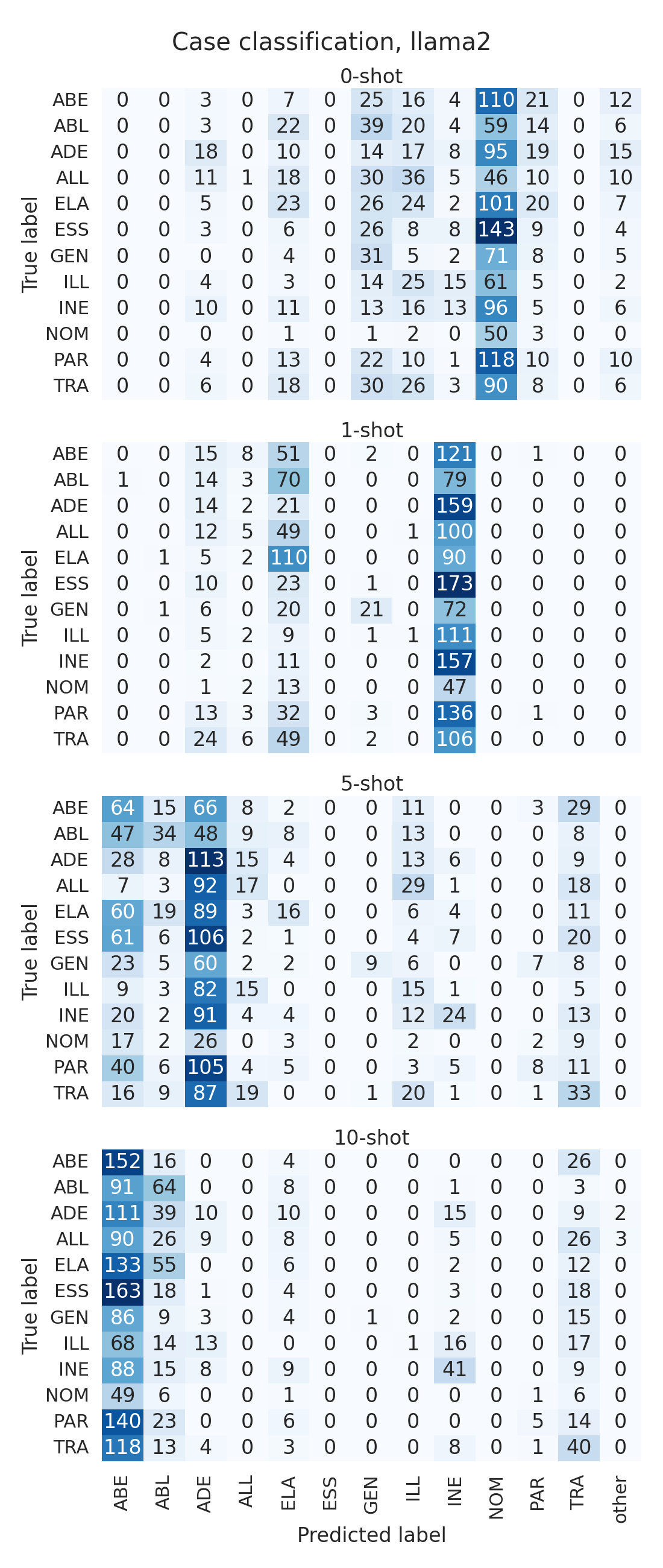}
    \caption{Confusions of Llama2-70B in the case classification task.}
    \label{fig:conf-llama-case}
\end{figure}

\begin{figure}[htb]
    \includegraphics[width=\columnwidth]{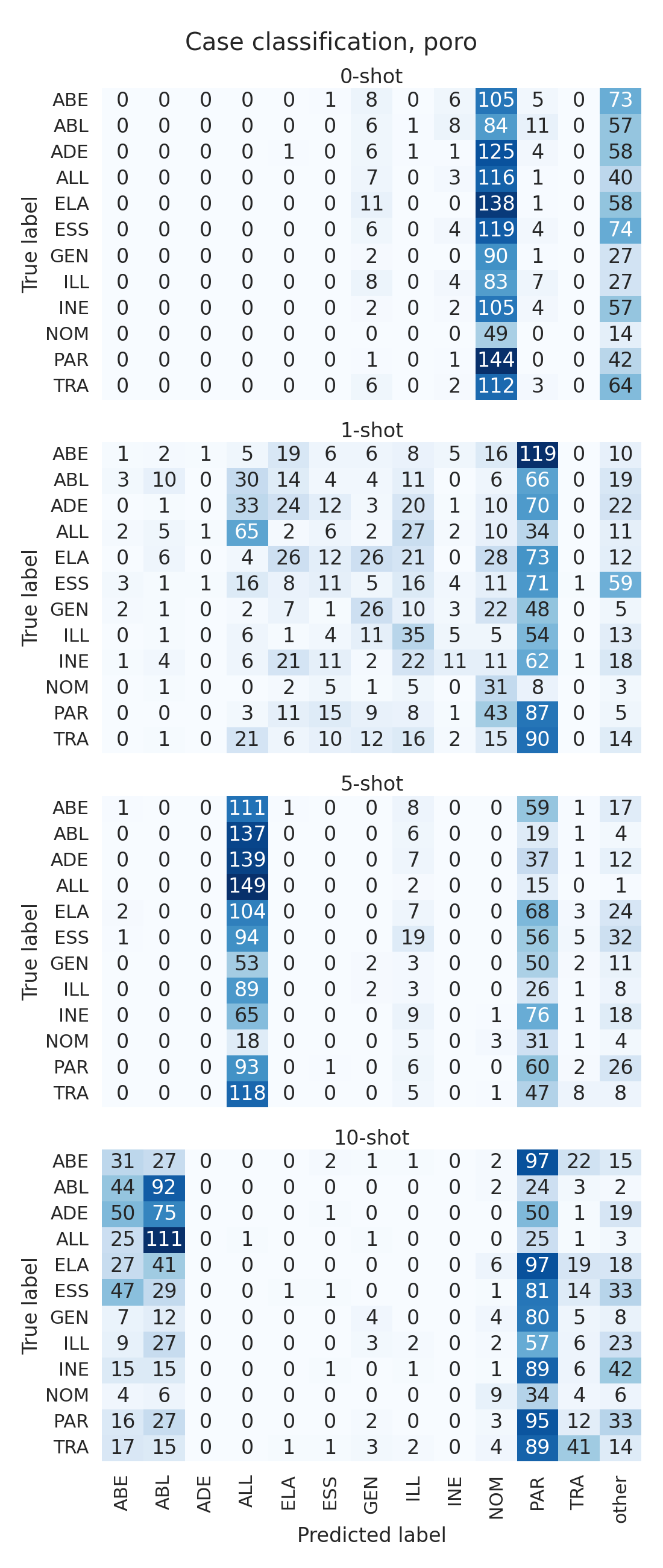}
    \caption{Confusions of Poro-34B in the case classification task.}
    \label{fig:conf-poro-case}
\end{figure}

\begin{figure}[htb]
    \includegraphics[width=0.9\columnwidth]{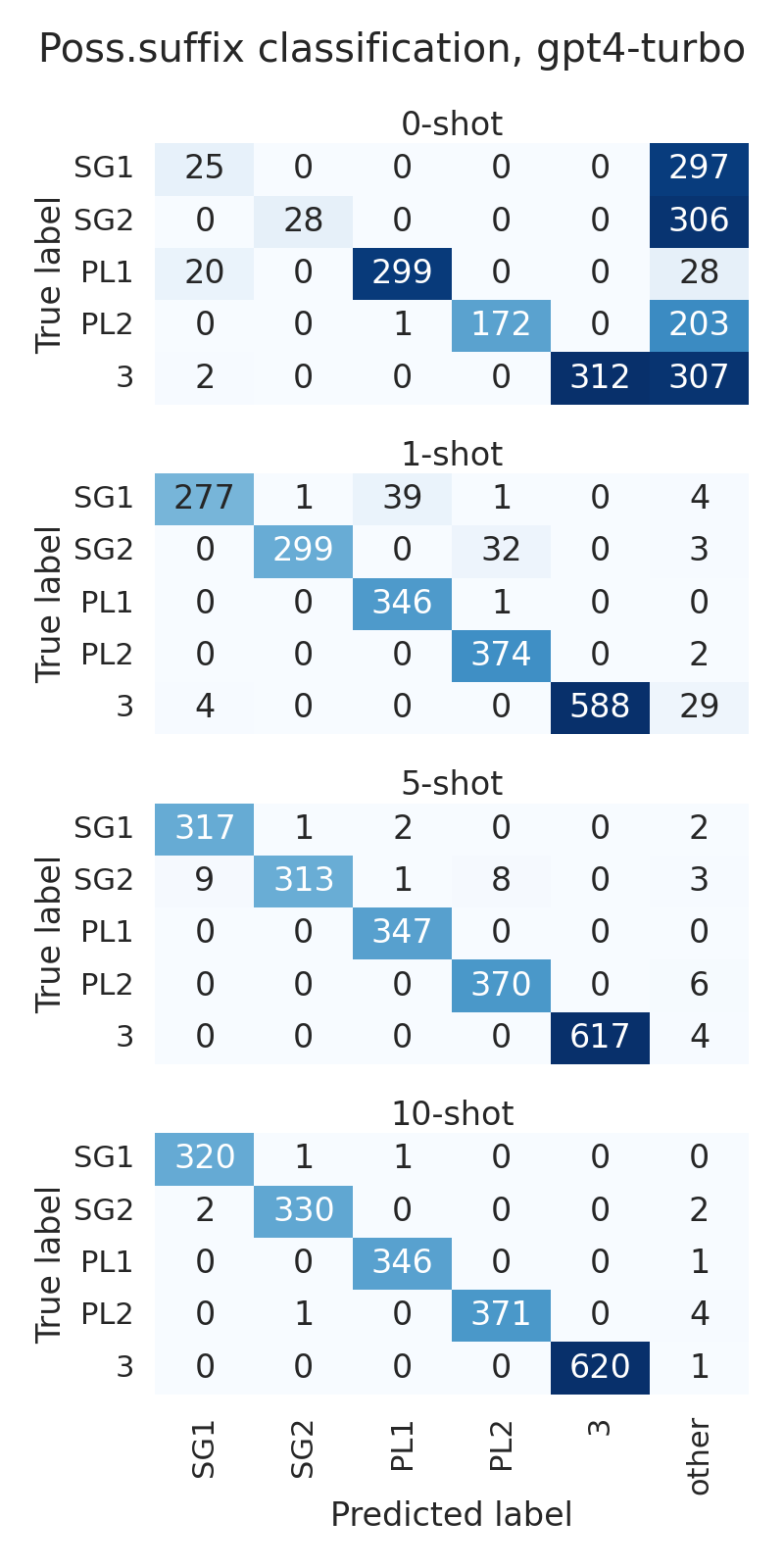}
    \caption{Confusions of GPT-4-turbo in the possessive suffix classification task.}
    \label{fig:conf-gpt4-person}
\end{figure}

\begin{figure}[htb]
    \includegraphics[width=0.9\columnwidth]{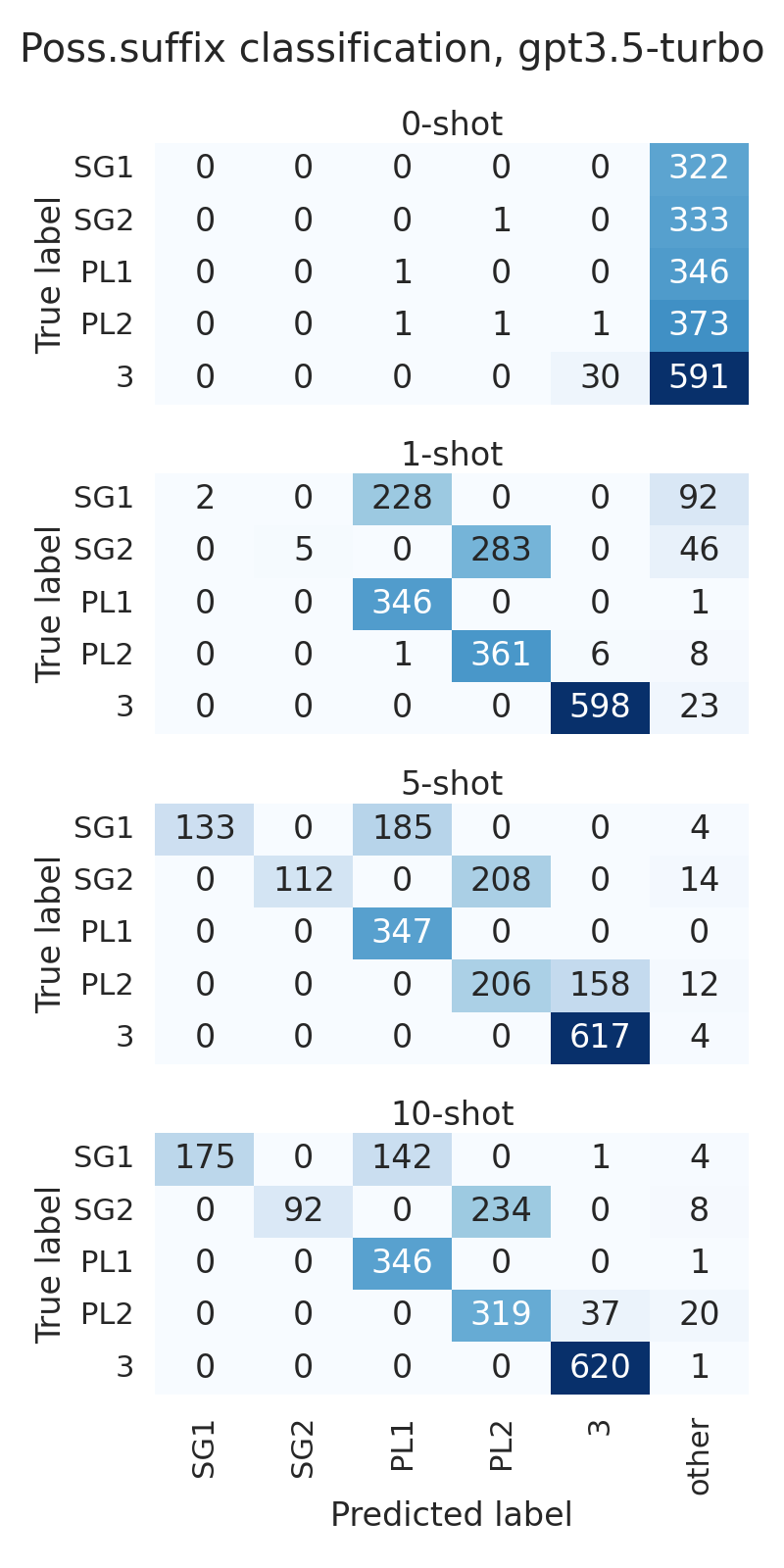}
    \caption{Confusions of GPT-3.5-turbo in the possessive suffix classification task.}
    \label{fig:conf-gpt35-person}
\end{figure}

\begin{figure}[htb]
    \includegraphics[width=0.9\columnwidth]{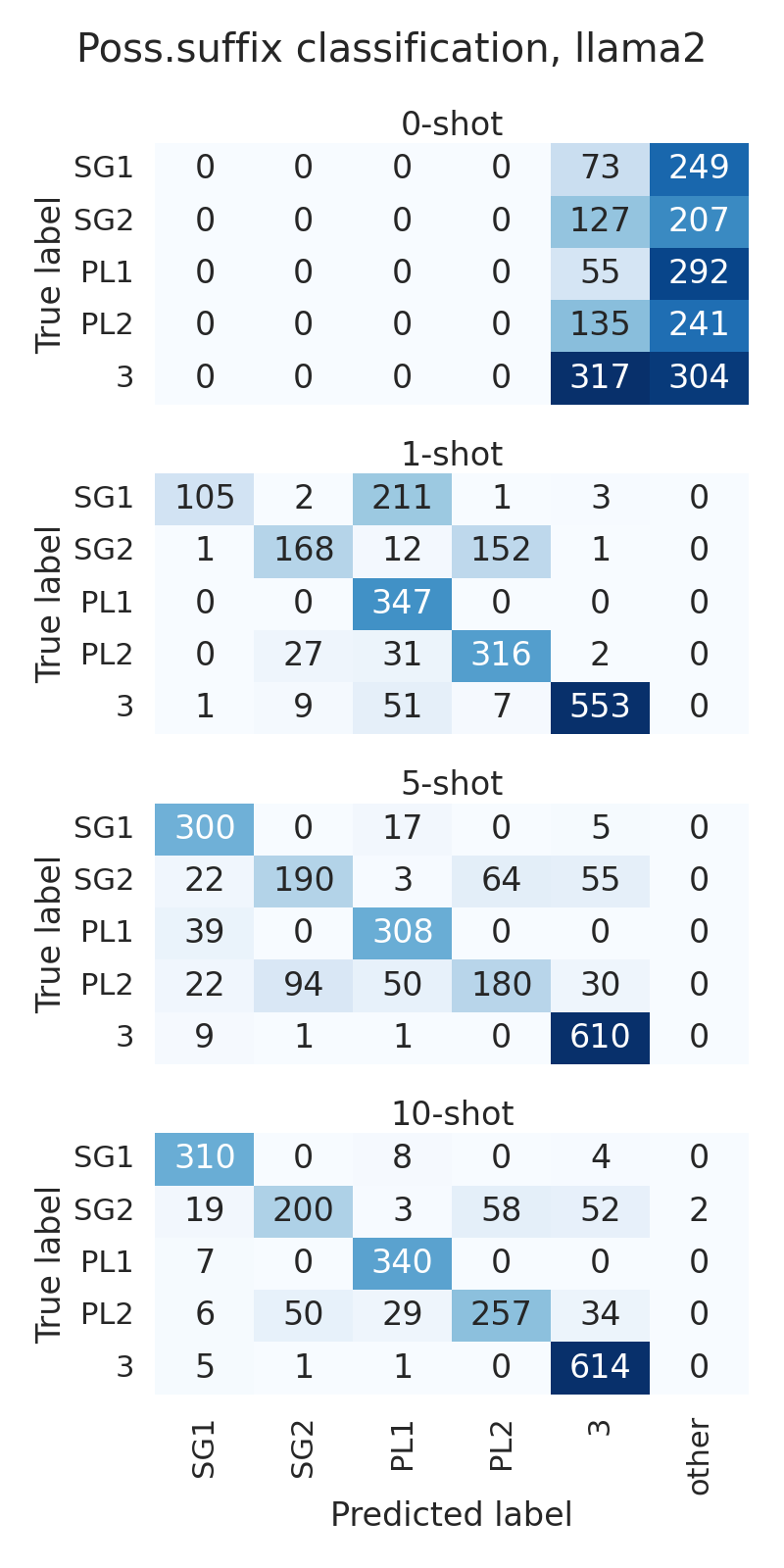}
    \caption{Confusions of Llama2-70B in the possessive suffix classification task.}
    \label{fig:conf-llama-person}
\end{figure}

\begin{figure}[htb]
    \includegraphics[width=0.9\columnwidth]{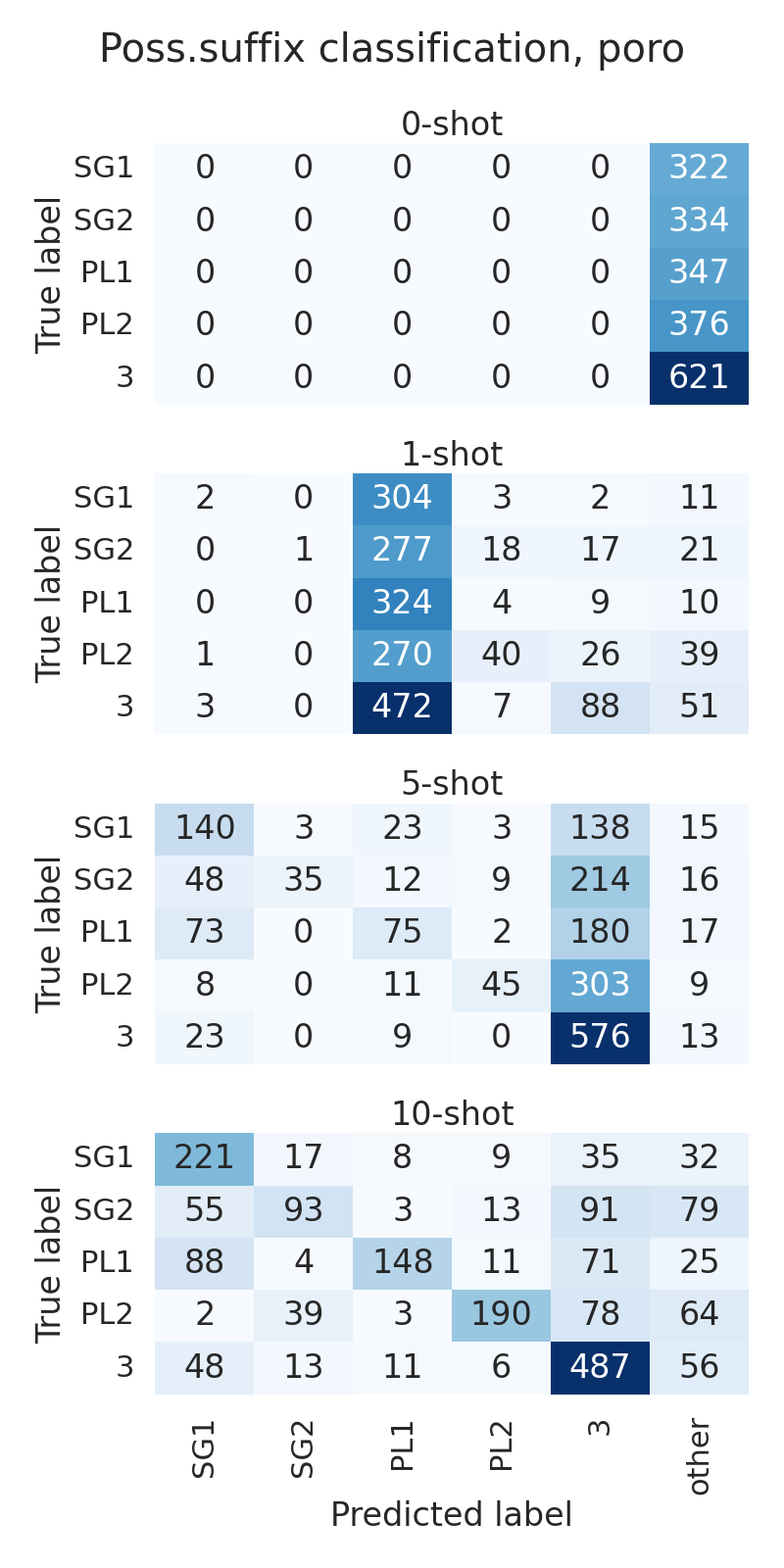}
    \caption{Confusions of Poro-34B in the possessive suffix classification task.}
    \label{fig:conf-poro-person}
\end{figure}

Figures~\ref{fig:num-gpt} through \ref{fig:conf-poro-person} show the confusion matrices of all models and in all classification tasks. Not all rows sum up to exactly to the same number: for example, in Figure~\ref{fig:num-llama-poro} 1-shot matrices, the SG row for Llama2 adds up to 964, whereas for Poro it adds up to 962. This is because of ambiguity in the task: for example the form `taloni' could be singular or plural (if the case is nominative). If the a system gives one of the correct classes, the `true label' is also assigned to that class in these confusion matrices. If the system predicts incorrectly, the `true label' could be any of the correct classes (whichever happens to be listed last in our data).

One notable thing in the confusion matrices is that Llama2-70B does not give many nonsense answers: when one or more examples are given in the prompt, the Llama2-70B almost always gives class names, correct or incorrect, which are actual classes, leaving the `other' column empty in Figures~\ref{fig:num-llama-poro}, \ref{fig:conf-llama-case}, and \ref{fig:conf-llama-person}. One reason that this is not the case for the GPT models is probably that GPT-4-turbo and GPT-3.5-turbo have been tuned for chat. In Microsoft Azure docs it is stated that `Like GPT-3.5 Turbo, and older GPT-4 models, GPT-4 Turbo is optimized for chat and works well for traditional completions tasks.'\footnote{\url{https://learn.microsoft.com/en-us/azure/ai-services/openai/concepts/models}}. GPT-4-turbo therefore often asks for clarification if it doesn't recognise the word, leading to nonsense classifications. Poro, on the other hand, is not tuned for chat, but still gives a lot of `other' answers. This seems to be more about Poro not grasping the format that the answer should be given in, or simply not knowing which classes are possible answers.




\end{document}